\documentclass[letterpaper, 10 pt, conference]{ieeeconf}  

\IEEEoverridecommandlockouts                             

\usepackage{cite}
\usepackage{amsmath,amssymb,amsfonts}
\usepackage{graphicx,import}
\usepackage[table]{xcolor}
\usepackage{siunitx}
\usepackage{svg}
\usepackage{graphicx}
\usepackage{caption}
\usepackage{subcaption}
\usepackage{algorithm}
\usepackage{algorithmic}
\def\BibTeX{{\rm B\kern-.05em{\sc i\kern-.025em b}\kern-.08em
		T\kern-.1667em\lower.7ex\hbox{E}\kern-.125emX}}
\usepackage{afterpage}
\usepackage{bm}
\newcommand{\vc}[1]{\bm{#1}}

\newcommand{\matb}[1]{\begin{bmatrix}#1\end{bmatrix}}

\title{\LARGE \bf Safety-Guaranteed Imitation Learning from Nonlinear Model Predictive Control for Spacecraft Close Proximity Operations}

\author{Alexander Meinert$^{1}$, Niklas Baldauf$^{1}$, Peter Stadler$^{1}$ and Alen Turnwald$^{2}$
\thanks{This work was supported by the Bavarian Ministry of Economic Affairs, Regional Development and Energy (grant no. MRF-2307-0009).}
\thanks{$^{1}$Alexander Meinert, Niklas Baldauf and Peter Stadler are with the Space Applications Group, e:fs TechHub GmbH, Gaimersheim, Germany. {\tt\small \{Alexander.Meinert, Niklas.Baldauf, Peter.Stadler\}@efs-techhub.com}}
\thanks{$^{2}$Alen Turnwald is with the Faculty of Electrical Engineering and Information Technology, Ingolstadt University of Applied Sciences, Germany. {\tt\small Alen.Turnwald@thi.de}}
}

\begin{document}
\maketitle
\thispagestyle{empty}
\pagestyle{empty}

\begin{abstract}
This paper presents a safety-guaranteed, runtime-efficient imitation learning framework for spacecraft close proximity control. We leverage Control Barrier Functions (CBFs) for safety certificates and Control Lyapunov Functions (CLFs) for stability as unified design principles across data generation, training, and deployment. First, a nonlinear Model Predictive Control (NMPC) expert enforces CBF constraints to provide safe reference trajectories. Second, we train a neural policy with a novel CBF-CLF-informed loss and {DA}\textsc{gger}-like rollouts with curriculum weighting, promoting data-efficiency and reducing future safety filter interventions. Third, at deployment a lightweight one-step CBF-CLF quadratic program minimally adjusts the learned control input to satisfy hard safety constraints while encouraging stability. We validate the approach for ESA-compliant close proximity operations, including fly-around with a spherical keep-out zone and final approach inside a conical approach corridor, using the Basilisk high-fidelity simulator with nonlinear dynamics and perturbations. Numerical experiments indicate stable convergence to decision points and strict adherence to safety under the filter, with task performance comparable to the NMPC expert while significantly reducing online computation. A runtime analysis demonstrates real-time feasibility on a commercial off-the-shelf processor, supporting onboard deployment for safety-critical on-orbit servicing.
\end{abstract}

\section{Introduction} \label{sec: Introduction}
With applications encompassing active debris removal, refueling, and in-orbit assembly, on-orbit servicing (OOS) is expected to play a central role in the next generation of space missions. Close proximity operations (CPO), in which a servicing spacecraft maneuvers near a target object, represent a safety-critical phase during an OOS mission. Given the substantial risks involved, space agencies, industry, and academia have developed comprehensive requirements for the design of safe CPO \cite{NASA2019, ISO243302022}, most recently exemplified by ESA's guidelines for the ClearSpace-1 mission \cite{ESA2024, Vasconcelos2025}.

Notably, these requirements pose complex challenges for the controller design with onboard autonomy considering safety zones, physical constraints, and the nature of nonlinear multi-input multi-output systems. Recent advances in Model Predictive Control (MPC) show promising results, as it enables the integration of the above challenges into a constrained optimal control problem (OCP). For example, MPC has been successfully applied to space applications including spacecraft rendezvous \cite{Rebollo2024} and attitude control \cite{Turnwald2023}.
However, the computational burden of solving nonlinear MPC (NMPC) problems on commercial off-the-shelf (COTS) space-grade processors has drawn attention to Imitation Learning (IL) as a runtime-efficient alternative, yet such data-driven controllers inherently lack safety and stability guarantees \cite{Reiter2026}.

A growing line of research---safe learning control \cite{Brunke2022}---investigates the joint combination of learning-based controllers with subsequent safety and stability filters to certify hard constraint satisfaction. The main idea behind safety filters is to verify whether a preceding, e.g. learning-based, control input keeps the system within a forward invariant safe set and, if necessary, to minimally modify the control input to ensure the trajectory adheres to safety constraints. Among well-established methods, Control Barrier Functions (CBF) have proven to be computationally lightweight as they can be implemented through a one-step quadratic program (QP) \cite{Wabersich2023}. Additionally, Control Lyapunov Functions (CLF) provide a means to enforce stability and can be combined with CBFs into a QP-based filter.

While several studies regarding safety in learning-based controllers have been conducted in the fields of robotics \cite{Mamedov2024} and autonomous driving \cite{Cosner2022}, only few approaches have been published in the context of spacecraft CPO. Relevant works include \cite{Kim2025}, where a long short-term memory neural network (NN) is integrated into a learning-based time shift governor scheme to enforce safety constraints during the final approach of spacecraft rendezvous; however, keep out zones are not considered in this framework. In \cite{Chen2023}, the authors train a reinforcement learning agent for trajectory planning in proximity operations, yet only incorporate line-of-sight constraints through a soft-constrained reward function, thus missing hard safety guarantees. Another body of literature \cite{Guffanti2024, Celestini2024} employs transformer-based neural networks to initialize trajectory optimization problems, thereby improving computational efficiency and preserving constraint satisfaction.

In contrast to the aforementioned prior work, our approach introduces two principal distinctions. First, we learn a direct state-to-control mapping, avoiding the need to predict entire MPC trajectories and thus eliminating the computational overhead of sequential models such as transformers. Second, and more fundamentally, our learning-based controller is deployed directly onboard, and paired with a CBF-CLF-filter to guarantee constraint satisfaction. Unlike existing methods \cite{Guffanti2024, Celestini2024} that use neural networks merely to warm-start an optimal control solver, our framework enables real-time execution with minimal computational and memory requirements—limited to a single neural network inference and a lightweight quadratic program, rather than repeatedly solving a non-convex OCP.

The present work is further motivated by evidence from \cite{Mammarella2025} and \cite{Drgoa2025}, which underscore the importance of safety certificates for spacecraft control in future missions, explicitly pointing out the demand for unified safety frameworks in learning-based settings. Briefly, our main contributions are:
\begin{itemize}
	\item A computationally efficient, learning-based control framework for spacecraft close proximity operations that understands safety as a unified design principle, incorporating CBF constraints across data generation, imitation learning training, and real-time deployment.
	\item A novel formulation of a CBF-CLF-informed loss function that integrates the safety and stability requirements of on-orbit servicing into data-efficient imitation learning.
	\item Robust validation of the control framework for the safety-critical rendezvous scenario---reflecting the requirements of the recently published ESA guidelines on safe CPO---using the high-fidelity astrodynamics simulator \textit{Basilisk}, including realistic orbital perturbations and nonlinear relative dynamics.
	\item Demonstration of real-time feasibility on embedded COTS hardware, supporting onboard deployment.
\end{itemize}
To the best of our knowledge, this is the first unified CBF-based safety-centric learning approach for spacecraft close proximity control, rigorously validated through simulation and real-time hardware demonstration.

The rest of this paper is structured as follows: Section \ref{sec: Problem Statement} presents the concept of operations relevant to the investigated CPO scenario. Section \ref{sec: Method} introduces the proposed imitation learning framework and provides details on the integration of safety and stability guarantees. The control framework is validated in simulation and through a runtime analysis on an embedded system in Section \ref{sec: Numerical Results}, and finally key findings and an outlook on future work are summarized in Section \ref{sec: Conclusion}.

\section{Concept of Operations} \label{sec: Problem Statement}
The concept of operations defines the mission-specific timeline and characteristics. We build upon the recently published ESA guidelines on safe CPO \cite{ESA2024} that were validated in \cite{Vasconcelos2025} with application to the ClearSpace-1 mission, where the aim is to capture and de-orbit an uncooperative target spacecraft. In this regard, the present work addresses the Close Rendezvous (CR) phase, in which the servicing spacecraft perceives the client with its onboard relative navigation sensors and performs active forced motion control. As an extension to the general-purpose ESA guidelines, for a given OOS scenario with a robotic arm mounted on the servicer, the CR consists of two operations.

\begin{figure}[h!]
	\setlength{\belowcaptionskip}{-15pt}
	\centering
	\def\svgwidth{0.75\linewidth}
\begingroup%
  \makeatletter%
  \providecommand\color[2][]{%
    \errmessage{(Inkscape) Color is used for the text in Inkscape, but the package 'color.sty' is not loaded}%
    \renewcommand\color[2][]{}%
  }%
  \providecommand\transparent[1]{%
    \errmessage{(Inkscape) Transparency is used (non-zero) for the text in Inkscape, but the package 'transparent.sty' is not loaded}%
    \renewcommand\transparent[1]{}%
  }%
  \providecommand\rotatebox[2]{#2}%
  \newcommand*\fsize{\dimexpr\f@size pt\relax}%
  \newcommand*\lineheight[1]{\fontsize{\fsize}{#1\fsize}\selectfont}%
  \ifx\svgwidth\undefined%
    \setlength{\unitlength}{465bp}%
    \ifx\svgscale\undefined%
      \relax%
    \else%
      \setlength{\unitlength}{\unitlength * \real{\svgscale}}%
    \fi%
  \else%
    \setlength{\unitlength}{\svgwidth}%
  \fi%
  \global\let\svgwidth\undefined%
  \global\let\svgscale\undefined%
  \makeatother%
  \begin{picture}(1,1.36134339)%
    \lineheight{1}%
    \setlength\tabcolsep{0pt}%
    \put(0,0){\includegraphics[width=\unitlength,page=1]{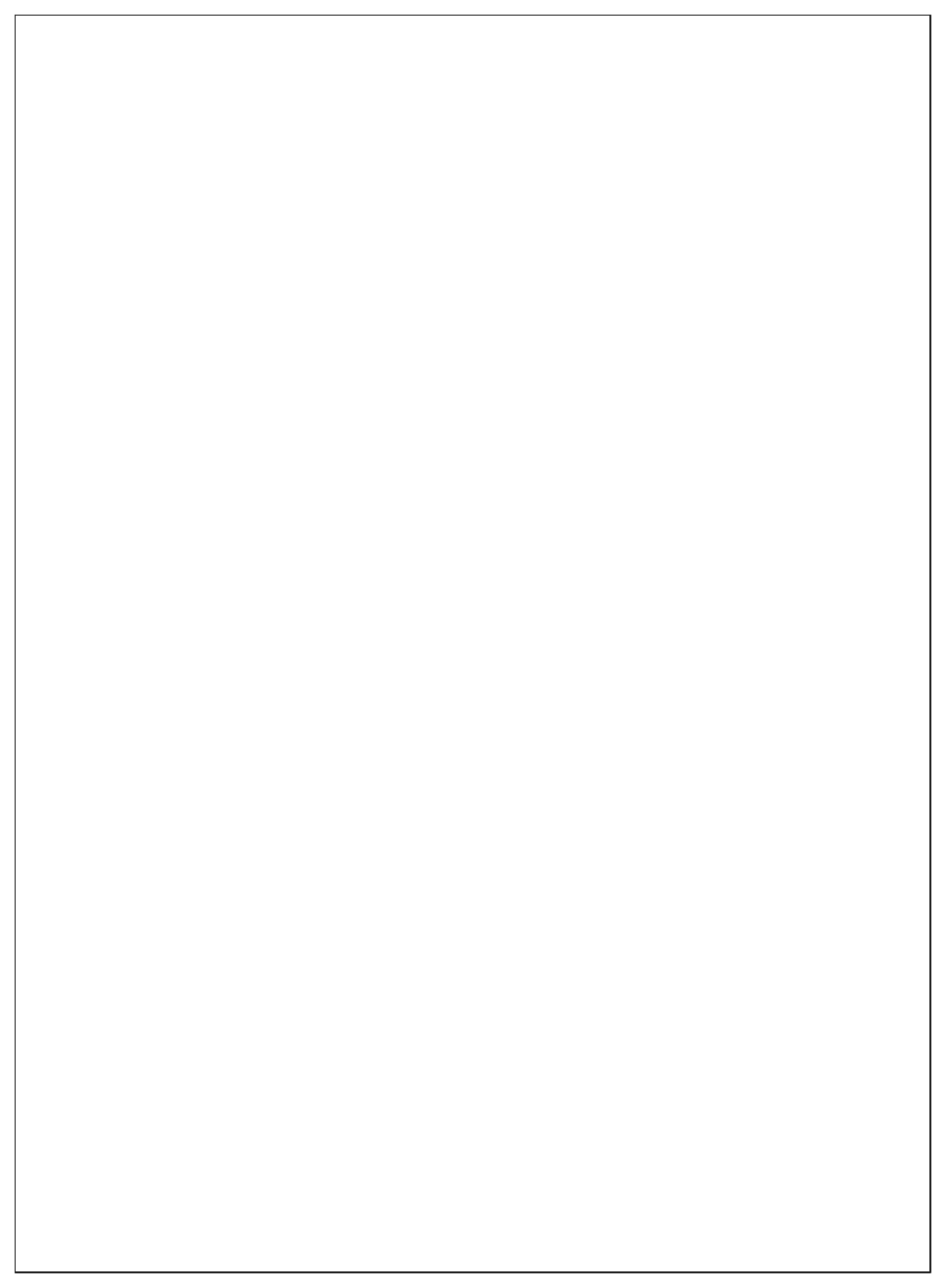}}%
    \put(0.47199554,1.29251374){\makebox(0,0)[lt]{\lineheight{1.25}\smash{\begin{tabular}[t]{l}$2\theta$\end{tabular}}}}%
    \put(0.71453976,0.4701858){\rotatebox{40.97834655}{\makebox(0,0)[lt]{\lineheight{1.25}\smash{\begin{tabular}[t]{l}$l_t$\end{tabular}}}}}%
    \put(0.2576928,0.20474993){\rotatebox{40.97834655}{\makebox(0,0)[lt]{\lineheight{1.25}\smash{\begin{tabular}[t]{l}$l_s$\end{tabular}}}}}%
    \put(0.78018514,0.40316454){\rotatebox{40.97834655}{\makebox(0,0)[lt]{\lineheight{1.25}\smash{\begin{tabular}[t]{l}$r_{KOZ}$\end{tabular}}}}}%
    \put(0,0){\includegraphics[width=\unitlength,page=2]{conops.pdf}}%
  \end{picture}%
\endgroup%

	\caption{Concept of operations for the Close Rendezvous mission phase in accordance with the ESA guidelines on safe CPO \cite{ESA2024}. The servicer performs two operations requiring forced motion control with active safety regards within the Approach Zone. (1) Spherical fly-around to decision point \textit{GO for KOZ} in order to align with the target angular momentum vector while avoiding the Keep Out Zone. (2) Final approach to decision point \textit{GO for Capture}, where entry in KOZ is only permitted if the conical Approach Corridor bounds are satisfied. The radius of the KOZ $r_{\text{KOZ}}$ is the sum of the largest dimensions $l_{\text{s}}$ and $l_{\text{t}}$ of the servicer and target, respectively.}
	\label{fig:conops}
\end{figure}

First, starting inside the Approach Zone (AZ), the servicer performs a spherical fly-around to reach the decision point \textit{GO for KOZ} while safely avoiding the Keep Out Zone (KOZ). The \textit{GO for KOZ} point serves as a designated hold point for station keeping since the servicer must synchronize its rotational motion before approaching the non-cooperative target along the angular momentum vector. The radius of the KOZ is chosen as the sum of the largest dimensions of the servicer's and target's cuboid clearance envelope, respectively, ensuring the minimum safety distance to avoid any potential collision. Second, the servicer enters the KOZ to reach the subsequent decision point \textit{GO for Capture} while staying inside the Approach Corridor (AC) aligned with the target angular momentum vector. Figure \ref{fig:conops} visualizes the concept of operations schematics.
Assuming the client is in a circular orbit, the Clohessy-Wiltshire-Hill equations of relative motion \cite{CLOHESSY1960} are expressed in the Local-Vertical Local-Horizontal (LVLH) reference frame in state-space form as follows
\begin{equation}\label{eq: state-space model}
	\small
	\setlength{\arraycolsep}{2.5pt}
	\renewcommand{\arraystretch}{1.0}
	\begin{bmatrix}
		\dot{x}_1 \\ \dot{x}_2 \\ \dot{x}_3 \\ \ddot{x}_1 \\ \ddot{x}_2 \\ \ddot{x}_3
	\end{bmatrix} = 
	\begin{bmatrix}
		0 & 0 & 0 & 1 & 0 & 0 \\
		0 & 0 & 0 & 0 & 1 & 0 \\
		0 & 0 & 0 & 0 & 0 & 1 \\
		3n^2 & 0 & 0 & 0 & 2n & 0 \\
		0 & 0 & 0 & -2n & 0 & 0 \\
		0 & 0 & -n^2 & 0 & 0 & 0 
	\end{bmatrix}
	\begin{bmatrix}
		x_1 \\ x_2 \\ x_3 \\ \dot{x}_1 \\ \dot{x}_2 \\ \dot{x}_3
	\end{bmatrix}
	+ 
	\begin{bmatrix}
		0 & 0 & 0 \\
		0 & 0 & 0 \\
		0 & 0 & 0 \\
		1 & 0 & 0 \\
		0 & 1 & 0 \\
		0 & 0 & 1
	\end{bmatrix}
	\begin{bmatrix}
		u_1 \\ u_2 \\ u_3
	\end{bmatrix}
	\normalsize
\end{equation}
where the states $\small\matb{x_1 & x_2 & x_3 & \dot{x}_1 & \dot{x}_2 & \dot{x}_3}^T$ are the relative distance and velocity components in R-bar, V-bar and H-bar direction, correspondingly, $n$ denotes the mean orbital motion, and $u_1, u_2, u_3$ are the input acceleration components. Note that we leverage the attitude controller from \cite{Meinert2024} to point the servicer onboard relative navigation sensors at the target throughout the entire CR phase.

\section{Method} \label{sec: Method}
\begin{figure*}[h]
	\setlength{\belowcaptionskip}{-15pt}
	\centering
	\def\svgwidth{\linewidth}
	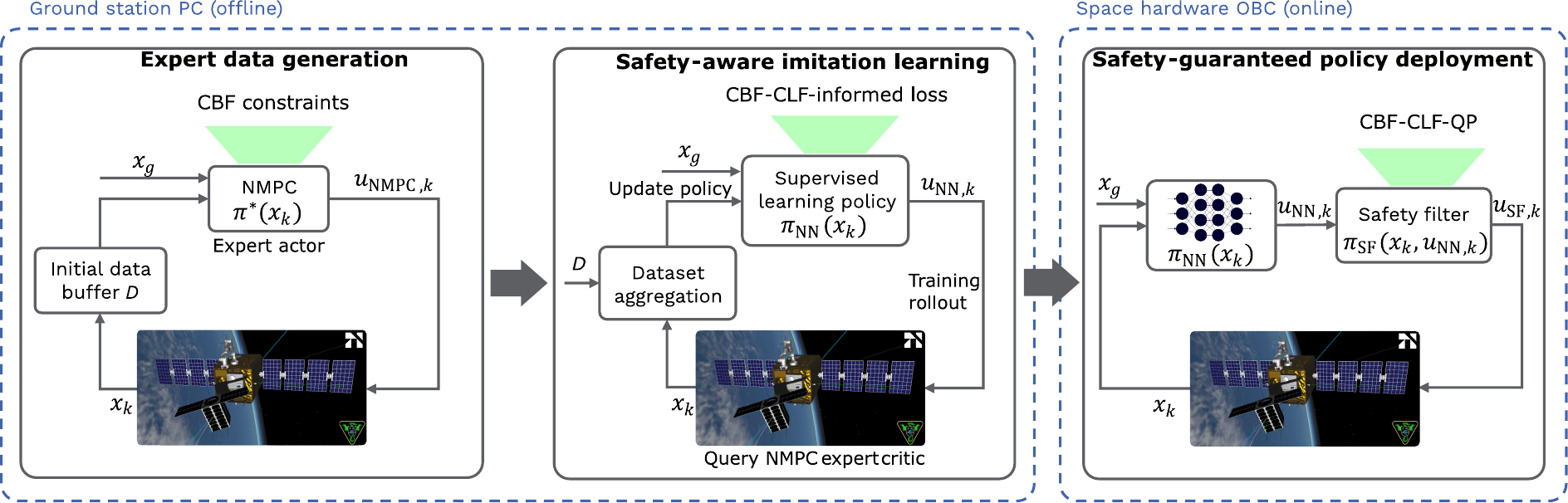
	\caption{Overview of the threefold, safety-guaranteed imitation learning framework, which incorporates safety considerations at multiple stages. (1) Data generation: Safety is integrated during data collection, as the NMPC expert policy enforces CBF constraints within the optimal control problem. (2) Training: A tailored loss function augments the conventional imitation loss with CBF and CLF soft constraints, imposing physical safety and stability priors into the learning process. (3) Deployment: During online deployment, a runtime-efficient CBF-CLF-QP safety filter ensures stability and hard constraint satisfaction.}
	\label{fig:imit-framework}
\end{figure*}

This section presents the main result of this work leading to the threefold, safe imitation learning framework, illustrated in Figure \ref{fig:imit-framework}. The following subsections focus on the essential elements; the CBF and CLF design, the NMPC expert policy as well as the safety-aware learning and deployment including a safety filter.  

\subsection{Input Constrained Control Barrier Function Design}
Consider the nonlinear control-affine system 
\begin{equation}\label{eq:control-affine system}
	\dot{\vc{x}} = \vc{f}(\vc{x}) + \vc{g}(\vc{x})\,\vc{u}
\end{equation}
locally Lipschitz continuous in $\vc{x}$, where $\vc{x} \in \mathcal{X} \subset \mathbb{R}^n$ denotes the state and $\vc{u} \in $ $\mathcal{U} \subset \mathbb{R}^m$ the control input. In accordance with \cite{Ames2019}, we define $\mathcal{S}$ as the forward invariant safe set
\begin{equation}
	\mathcal{S} = \{\vc{x} \in \mathbb{R}^n\,|\, h(\vc{x}) \geq 0\}
\end{equation}
where $\mathcal{S}$ is the zero-superlevel set of the continuously differentiable constraint function $h(\vc{x})$. We refer to $h(\vc{x})$ as a Control Barrier Function if there exists an extended class-$\mathcal{K}$ function $\alpha \in \mathcal{K}$ such that the safety condition
\begin{equation} \label{eq:CBF rel deg 1}
	\sup\limits_{u \in \mathcal{U}} [L_fh(\vc{x}) + L_gh(\vc{x}) \vc{u}] \geq -\alpha(h(\vc{x}))
\end{equation}
holds $\forall \vc{x} \in \mathcal{S}$.
Safety is guaranteed by keeping the state trajectory inside the forward invariant safe set for any initial state $\vc{x}(0) \in \mathcal{S}$, since \eqref{eq:CBF rel deg 1} ensures that the time derivative $\dot{h}(\vc{x},\vc{u})$ is positive whenever $h(\vc{x})=0$, thereby driving the state back into $\mathcal{S}$.
However, satisfying the safety condition \eqref{eq:CBF rel deg 1} and actuator constraints at the same time might lead to infeasibility in a NMPC or safety filter QP. Thus, for the inertia-based second-order dynamics \eqref{eq: state-space model}, we leverage the results from \cite{Breeden2023} and formulate the input constrained CBF condition
\begin{equation} \label{eq:CBF rel deg 2}
	\sup\limits_{u \in \mathcal{U}} \mathbb{H}(\vc{x},\vc{u}) = \sup\limits_{u \in \mathcal{U}} \left[L_f H(\vc{x}) + L_g H(\vc{x}) \vc{u} + \alpha(H(\vc{x}))\right] \geq 0
\end{equation}
for a constraint function $h(\vc{x})$ with relative degree $\rho=2$ and $L_gh(\vc{x})=0$, where
\begin{align}
	H(\vc{x}) &= h(\vc{x}) + \frac{| \dot{h}(\vc{x}) | \dot{h}(\vc{x})}{2u_{\max}}\\
	\dot{H}(\vc{x},\vc{u}) &= \dot{h}(\vc{x}) + \frac{| \dot{h}(\vc{x}) | \ddot{h}(\vc{x},\vc{u})}{u_{\max}}.
\end{align}
In fact, \eqref{eq:CBF rel deg 2} implies the existence of an inner safe set $\mathcal{S}_H \subset \mathcal{S}$ that accounts for the simultaneous viability of safety and input constraints. We construct the following two CBFs for the CPO scenario outlined in Section \ref{sec: Problem Statement}.

\subsubsection{Spherical CBF for Keep Out Zone}
The spherical KOZ safety constraint for the fly-around is characterized by
\begin{equation} \label{eq:CBF for KOZ}
	h_1(\vc{x}) = \|\vc{p} - \vc{p}_o\|^2 - r_{\text{KOZ}}^2 \geq 0,
\end{equation}
in which $r_{\text{KOZ}}$ is the radius of the KOZ, $\small\vc{p}=\matb{x_1&x_2&x_3}^T$, and $\small\vc{p}_o=\matb{x_o&y_o&z_o}^T$ represents the center of the KOZ aligned with the origin of the LVLH frame for the given scenario.

\subsubsection{Conical CBF for Approach Corridor}
For the final approach, we construct a conical safety constraint
\begin{equation} \label{eq:CBF for AC}
	h_{2}(\vc{x}) = \frac{\vc{p}^T\,\vc{d}}{\lVert\vc{p}\rVert\,\lVert\vc{d}\rVert} - \cos(\theta) \geq 0,
\end{equation}
where $\vc{d}$ is the cone axis direction aligned with the target angular momentum vector and $\theta$ denotes the half-cone angle. For both CBFs \eqref{eq:CBF for KOZ} and \eqref{eq:CBF for AC} expressed in the form of \eqref{eq:CBF rel deg 2}, we set $\alpha(H(\vc{x}))=\gamma H(\vc{x})$, where $\gamma$ is a design parameter indicating the notion how fast the state trajectory is allowed to approach the boundary of the safe set.

\subsection{Control Lyapunov Function Design}
Another desirable property is to certify the stabilizability of the control-affine system \eqref{eq:control-affine system} to an equilibrium point $\vc{x}_g$, e.g. a decision point in CPO. For that matter, a positive definite function $V(\vc{x})$ is called a Control Lyapunov Function if, for every state $\vc{x} \neq \vc{x}_g$, there exists a control input such that
\begin{equation} \label{eq:CLF}
	\inf\limits_{u \in \mathcal{U}} [L_fV(\vc{x}) + L_gV(\vc{x}) \vc{u}] \leq -\beta(V(\vc{x}))
\end{equation}
for some class $\mathcal{K}$ function $\beta$ \cite{Ames2019}. This condition implies that each sublevel set of $V(\vc{x})$ can be rendered forward invariant, leading to stabilizability of $\vc{x}_g$. For the proximity control problem, a quadratic CLF with relative degree 1 is designed as
\begin{equation} \label{eq:V}
	V(\vc{x}) = (\vc{x}-\vc{x}_{g})^T\vc{P}(\vc{x}-\vc{x}_{g}),
\end{equation}
where $\vc{P}$ is the solution to the discrete-time algebraic Riccati equation. Moreover, preliminary experiments revealed that the imitation learner encounters difficulties in achieving stationary accuracy at the decision points. In order to enforce stability in close-vicinity to $\vc{x}_g$ through a safety filter---without interfering with the CBF safety constraints when far away from the decision point---\eqref{eq:CLF} is rewritten such that
\vspace{-8pt}

{\small
	\begin{equation}\label{eq:state-dependent CLF}
		\inf\limits_{u \in \mathcal{U}} \mathbb{V}(\vc{x},\vc{u}) = \inf\limits_{u \in \mathcal{U}} [L_fV(\vc{x}) + L_gV(\vc{x}) \vc{u} + \zeta(\vc{x})V(\vc{x})] \leq 0
	\end{equation}
}
with the state-dependent decay rate
\begin{equation} \label{eq:sigmoid}
	\zeta(\vc{x}) = \zeta_{\min} + (\zeta_{\max} - \zeta_{\min})\frac{1}{1 + e^{j\left(\|\vc{x} - \vc{x}_g\| - c\right)}}
\end{equation}
where $j$ and $c$ adjust the steepness and midpoint of the sigmoid function, respectively, and $\zeta_{\min}$ and $\zeta_{\max}$ represent the lower and upper bounds.

\subsection{Nonlinear Model Predictive Control for Spacecraft CPO}
A safe NMPC policy is employed to generate an expert dataset $\mathcal{D}$ used in the subsequent training of an imitation learner. The constrained finite time OCP, repeatedly solved at every time step via the IPOPT\footnote{https://github.com/coin-or/Ipopt} solver, yields
\vspace{-8pt}

\begin{subequations}\label{eq: MPC CFTOC}
	\small
	\begin{align}
		\vc{U}^{*}_{0}(\vc{x}(t)) =\, &\underset{\vc{U}_{0}}{\arg\!\min} \: \lVert \bar{\vc{x}}_N \rVert^2_{\vc{P}} 
		+\sum_{k = 0}^{N-1}\lVert \bar{\vc{x}}_k \rVert^2_{\vc{Q}} + \lVert \vc{u}_k \rVert^2_{\vc{R}} \label{eq:MPC cost}\\
		\text{s.t. } &\vc{x}_{k+1} = \vc{A}\,\vc{x}_{k} + \vc{B}\,\vc{u}_{k} \label{eq: MPC dynamics}\\
		& \vc{x}_k \in \mathcal{X}, \,\vc{u}_k \in \mathcal{U}, k=0,...,N-1 \label{eq: MPC box constraints}\\
		& \vc{x}_N \in \mathcal{X}_f \label{eq: terminal state constraints}\\
		& \vc{x}_0 = \vc{x}(t) \label{eq: initial state}\\
		& L_f H(\vc{x}_k) + L_g H(\vc{x}_k) \vc{u}_k + \gamma H(\vc{x}_k) \geq \epsilon \label{eq: MPC CBF constraint}
	\end{align}
	\normalsize
\end{subequations}
where \eqref{eq:MPC cost} contains the quadratic cost function with the tracking error $\bar{\vc{x}}=\vc{x}-\vc{x}_g$ and the weighting matrices $\vc{Q}\!=\!\vc{Q}^T\!\succeq 0$, $\vc{P}\!=\!\vc{P}^T\!\succeq 0$, and $\vc{R}\!=\!\vc{R}^T\!\succ 0$. The discretized system dynamics from \eqref{eq: state-space model} are given in \eqref{eq: MPC dynamics}, the state and input box constraints are defined by \eqref{eq: MPC box constraints}, the terminal state constraint is given in \eqref{eq: terminal state constraints}, and the current state is used to initialize the predicted trajectory by \eqref{eq: initial state}. Safety is ensured by the CBF constraint in \eqref{eq: MPC CBF constraint}, while $\epsilon$ represents a discretization offset to the safe set boundary. Note that enforcing the CBF constraint---either based on the spherical KOZ or conical AC depending on the operation during the CR---results in a computationally expensive nonlinear program. The solution to \eqref{eq: MPC CFTOC} is the optimal input sequence $\vc{U}^{*}_{0}$, while only the first element is applied as the control input $\vc{u}_{\text{NMPC}}(t) = \vc{u}^{*}_{0}(\vc{x}(t))$ to the system.

\subsection{CBF-CLF-Informed Loss Function}
Motivated by the objective to achieve a runtime-efficient controller, the expert policy in \eqref{eq: MPC CFTOC} is approximated using a supervised imitation learning strategy. While offline Behavior Cloning (BC), solely subject to the conventional imitation loss
\begin{equation} \label{eq:imit loss}
	\mathcal{L}_{\text{imit}} =	\mathbb{E}_{\vc{x} \sim \mathcal{D}} \left[ \left\lVert \pi_{\text{NN}}(\vc{x}) - \pi^*(\vc{x}) \right\rVert^2 \right],
\end{equation}
is prone to covariate shift---since the training data is sampled in areas where the expert policy leads to---we apply the on-policy {DA}\textsc{gger} \cite{Ross2011} algorithm. Here, a mixed policy consisting of the supervised trained NN and the expert is rolled out in simulation to gather new training data by evaluating the NMPC expert critic. At every {DA}\textsc{gger} iteration, the rollout policy's weighting $\kappa$ is shifted to rely more on the imitation-learned NN policy. In addition, we leverage the intent from physics-informed NN \cite{Drgoa2025} and formulate a novel CBF-CLF-informed loss function \eqref{eq:total loss} tailored to spacecraft close proximity control, hence incorporating the physical constraint knowledge into the training process. To do so, after an initial training phase based on the imitation loss, \eqref{eq:imit loss} is augmented by a CBF-specific loss
\begin{equation} \label{eq:CBF loss}
	\mathcal{L}_{\text{CBF}} = \mathbb{E}_{\vc{x} \sim D}\left[\max(0, -\mathbb{H}(\vc{x}, \pi_{\text{NN}}(\vc{x})))^2 \right]
\end{equation}
and a CLF-informed loss term
\begin{equation} \label{eq:CLF loss}
	\mathcal{L}_{\text{CLF}} = \mathbb{E}_{\vc{x} \sim D}\left[\max(0, \mathbb{V}(\vc{x}, \pi_{\text{NN}}(\vc{x})))^2 \right]
\end{equation}
that penalize the NN only if the CBF \eqref{eq:CBF rel deg 2} or CLF \eqref{eq:state-dependent CLF} constraints are violated. Notably, curriculum learning is performed, increasing the weights $\lambda_{\text{CBF}}$ and $\lambda_{\text{CLF}}$ towards the soft-constrained safety \eqref{eq:CBF loss} and stability \eqref{eq:CLF loss} aware loss terms, accordingly, where the total loss function over the course of the {DA}\textsc{gger} refinement culminates in
\begin{equation} \label{eq:total loss}
	\mathcal{L}_{\text{total}} = \lambda_{\text{imit}}\,\mathcal{L}_{\text{imit}} + \lambda_{\text{CBF}}\,\mathcal{L}_{\text{CBF}} + \lambda_{\text{CLF}}\,\mathcal{L}_{\text{CLF}}.
\end{equation}
By introducing the loss function in \eqref{eq:total loss}, we aim at two goals; namely, (i) data-efficiency through less required data samples; (ii) minimal safety filter activation during the deployment stage. Algorithm \ref{alg:IL} outlines the training procedure and parameter update rules. 
\setlength{\textfloatsep}{10pt}
\begin{algorithm}[t!]
\caption{Safety-aware imitation learning from NMPC}
\label{alg:IL}
\begin{algorithmic}[1]
	\renewcommand{\algorithmicrequire}{\textbf{Input:}}
	\renewcommand{\algorithmicensure}{\textbf{Output:}}
	\REQUIRE $\mathcal{D}$, $\lambda_{\text{imit}}$, $n_{\text{epochs}}$, $n_{\text{iter}}$, $n_{\text{rollout}}$
	\ENSURE Trained policy $\pi_{\text{NN}}$
	\FOR{$i = 1$ to $n_{\text{epochs}}$}
		\FOR{each $\text{batch}$ in $\mathcal{D}$}
			\STATE Train $\pi_{\text{NN}}$ using imitation loss~\eqref{eq:imit loss}
		\ENDFOR
	\ENDFOR
	\STATE Initialize $\lambda_{\text{CBF}}$, $\lambda_{\text{CLF}}$
	\FOR{$i = 1$ \textbf{to} $n_{\text{iter}}$} 
		\STATE Update $\kappa \gets 1 - i / n_{\text{iter}}$
		\FOR{$j = 1$ \textbf{to} $n_{\text{rollout}}$}
			\STATE Infer imitation policy $\pi_{\text{NN}}(\vc{x})$
			\STATE Query NMPC expert $\pi^*(\vc{x})$ \eqref{eq: MPC CFTOC}
			\STATE Compute $\pi_{\text{rollout}}(\vc{x}) = \kappa \pi^*(\vc{x}) + (1 - \kappa) \pi_{\text{NN}}(\vc{x})$
			\STATE Apply $\pi_{\text{rollout}}(\vc{x})$: $\vc{x} \gets \text{Simulate}(\vc{x}, \pi_{\text{rollout}}(\vc{x}))$
			\STATE Aggregate $\mathcal{D} \gets \mathcal{D} \cup (\vc{x}, \pi^*(\vc{x}))$
		\ENDFOR
		\FOR{each $\text{batch}$ in $\mathcal{D}$}
			\STATE Train $\pi_{\text{NN}}$ using CBF-CLF-informed loss~\eqref{eq:total loss}
		\ENDFOR
		\STATE Update $\lambda_{\text{CBF}} \gets 2\lambda_{\text{CBF}}$, $\lambda_{\text{CLF}} \gets 2\lambda_{\text{CLF}}$
	\ENDFOR
	\STATE \textbf{Return} $\pi_{\text{NN}}$
\end{algorithmic}
\end{algorithm}

\subsection{Safety Filter Design}
During the online deployment stage, hard safety guarantees are enforced through a minimally intrusive CBF-CLF-QP safety filter inspired by \cite{Ames2019} that adjusts the learning-based control input such that the state trajectory is kept inside the forward invariant safe set. In contrast to the NMPC problem in \eqref{eq: MPC CFTOC}, the safety filter
\begin{subequations}\label{eq: safety filter}
	\begin{align}
		\underset{\vc{u}_{\text{SF},k},\delta_k} \min &\lVert \vc{u}_{\text{SF},k}-\vc{u}_{\text{NN},k} \rVert^2 + p\,\delta_k^2 \label{eq:SF cost}\\
		\text{s.t. } & \mathbb{H}(\vc{x}, \vc{u}_{\text{SF},k}) \geq \epsilon \label{eq:SF CBF}\\
		& \mathbb{V}(\vc{x}, \vc{u}_{\text{SF},k}) \leq \delta_k \label{eq:SF CLF}\\
		& {\vc{u}_{\text{SF},k} \in \mathcal{U}} \label{eq:SF input constraints}
	\end{align}
\end{subequations}
can be solved as a convex, one-step QP with limited computational cost. Besides the quadratic objective function \eqref{eq:SF cost}, the CBF safety constraint \eqref{eq:SF CBF}, the CLF stability constraint \eqref{eq:SF CLF}, and the input constraints \eqref{eq:SF input constraints} are all linear in the decision variables. Note that the slack variable $\delta$ is introduced to ensure feasibility of \eqref{eq: safety filter} by prioritizing safety over stability; however, the state-dependent CLF decay rate in \eqref{eq:sigmoid} is tuned considering less conflicts of the constraints.
\setlength{\textfloatsep}{10pt}

\section{Numerical Results} \label{sec: Numerical Results}

\subsection{Scenario Setup}
For validation purposes, the control framework is integrated into the high-fidelity astrodynamics simulator $\textit{Basilisk}$ \cite{Kenneally2020} that accommodates orbital perturbations and nonlinearities. The circular client's orbit is at an altitude of $400\,\unit{km}$, a spherical AZ with radius $40\,\unit{m}$, $r_{\text{KOZ}}\!=\!10\,\unit{m}$, $\theta\!=\!3\unit{\degree}$ are given. The decision points are set as $15\,\unit{m}$ and $2.3\,\unit{m}$ in V-bar direction relative to the target center of mass for both operations, respectively, while a safety distance of $2\,\unit{m}$ must be maintained. The actuator input is bounded by $-0.082 \leq \vc{u} \leq 0.082\,\unit{m.s^{-2}}$ and $T_s\!=\!0.1\,\unit{s}$ denotes the sample time. System parameters are given as $\vc{Q}\!=\!\vc{I}_6$, $\vc{R}\!=\!10^4\,\vc{I}_3$, $N\!=\!10$,  $\gamma_1\!=\!0.5$, $\gamma_2\!=\!1$, $\epsilon\!=\!0.01$, $p\!=\!0.001$, $j\!=\!1$, $c\!=\!15$, $\zeta_{\min}\!=\!0.001$, $\zeta_{\max}\!=\!0.06$. A fully-connected feedforward NN with 4 layers, 256 neurons each and ReLU activation functions is trained using a learning rate of $10^{-4}$, the \textit{AdamW} optimizer, a batch size of 128, $n_{\text{epochs}}\!=\!20$, $n_{\text{iter}}\!=\!5$, $n_{\text{rollout}}\!=\!5$ and $81,027$ initial training samples, while $\lambda_{\text{imit}}\!=\!100$, $\lambda_{\text{CBF}}\!=\!10^{-7}$, $\lambda_{\text{CLF}}\!=\!10^{-5}$ are designed to yield losses of similar order of magnitude. The layers are normalized before activation to stabilize the training, a dropout rate of $0.1$ is employed to avoid overfitting, and gradients are clipped to $\pm 0.5$.

\subsection{Controller Ablation Study}
\begin{figure}[t!]
	\centering
	\begin{subfigure}[h]{0.9\linewidth}
		\centering
		\scalebox{0.27}{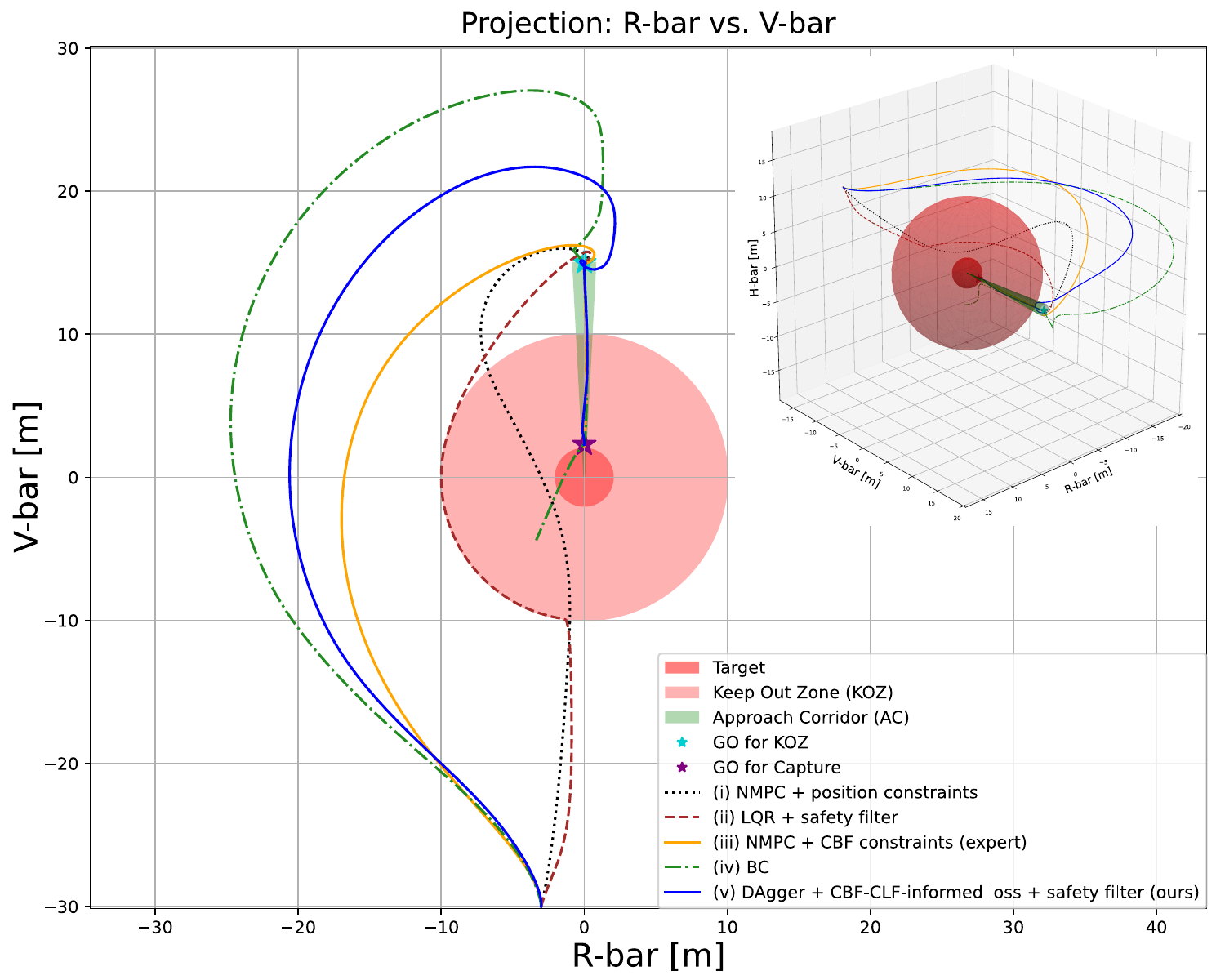}
		\caption{Comparison of closed-loop trajectories.}
		\label{fig:subfig2}
	\end{subfigure}
		\vspace{1pt}
	\begin{subfigure}[h]{0.9\linewidth}
		\scalebox{0.3}{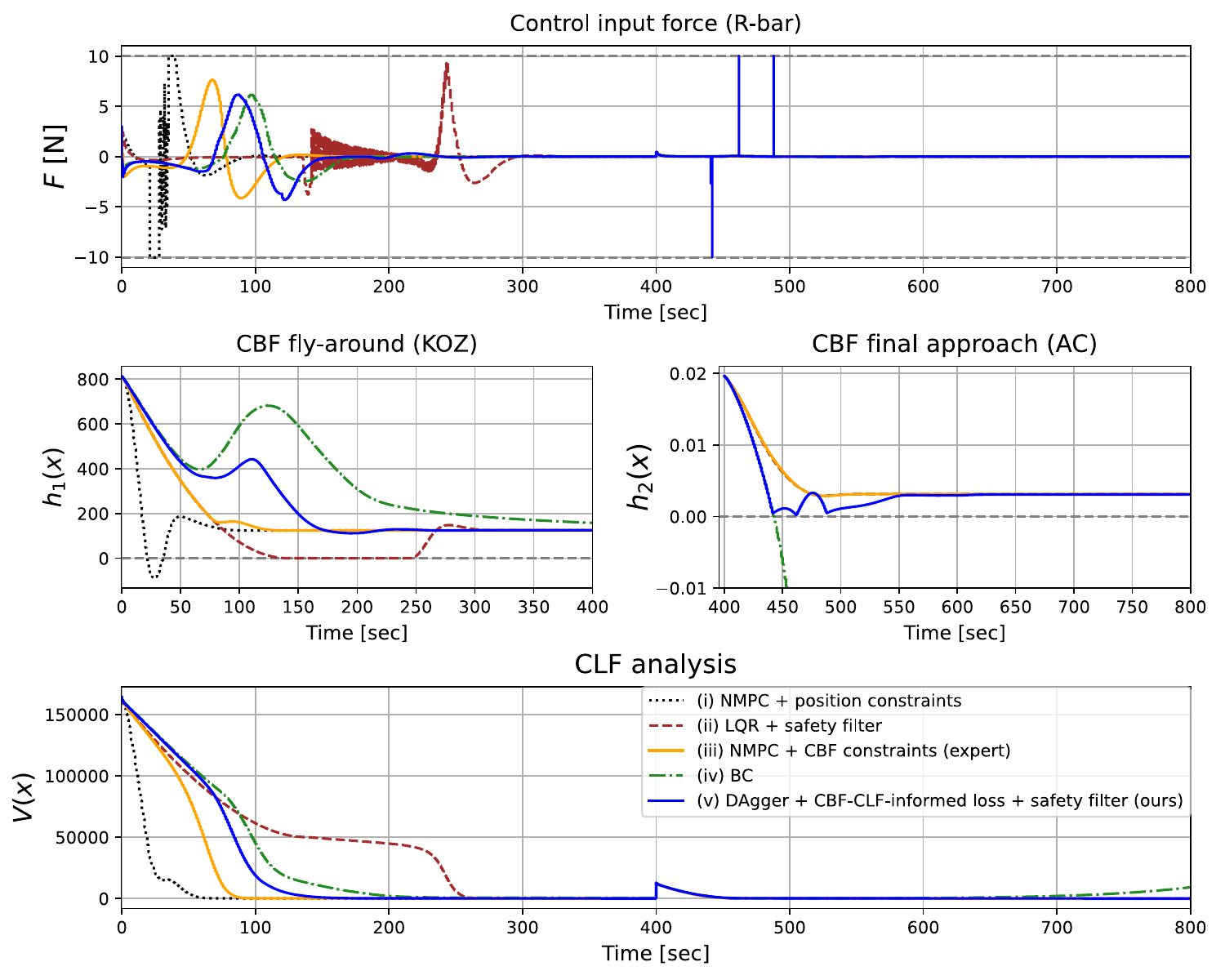}
		\caption{Control input and CBF-CLF analysis.}
		\label{fig:subfig3}
	\end{subfigure}
	\caption{Simulation results and controller ablation study.}
	\label{fig:results}
\end{figure}
An ablation study compares five control policies, namely, (i) NMPC with position avoidance constraints using \eqref{eq:CBF for KOZ} instead of \eqref{eq: MPC CBF constraint}; (ii) LQR with safety filter; (iii) expert NMPC with integrated CBF constraints \eqref{eq: MPC CFTOC}; (iv) conventional BC subject to pure imitation loss \eqref{eq:imit loss}; (v) our proposed control framework. Figure \ref{fig:results} shows the simulation results starting from $\small\vc{x}=\matb{-3 & -30 & 2 & 0 & 0 & 0}^T$ and Table \ref{tab:runtime} compares the runtime, including the embedded implementation on an ESP32-S3-N16R8 microcontroller (Xtensa dual-core LX7, 240 MHz) with a setup analogous to \cite{Turnwald2025}. Policy (i) becomes infeasible when satisfying the KOZ constraint and the input constraints at the same time, highlighting the need for relative degree 2 CBF constraints. The filter in (ii) guarantees safety, while yielding a primitive trajectory along the safe set boundary with vast filter interventions and input chattering, as the unconstrained LQR has no knowledge of the KOZ or AC. The expert (iii) provides an optimal trajectory regarding safety and input constraints according to \eqref{eq: MPC CFTOC}. BC (iv) lacks stable convergence to decision points and violates the AC constraint. On the contrary, (v) provides a safe and stabilizing imitation of the expert policy. In terms of task performance, it follows a slightly longer trajectory; however, the substantial reduction in execution time renders this trade-off worthwhile. Table \ref{tab:runtime} reveals that the NMPC (iii) exceeds the sample time in simulation, while the proposed controller (v) is tractable for both simulation and onboard deployment.

\begin{figure}[h!]
	\setlength{\belowcaptionskip}{-10pt}
	\def\svgwidth{0.90\linewidth}
	\centering
	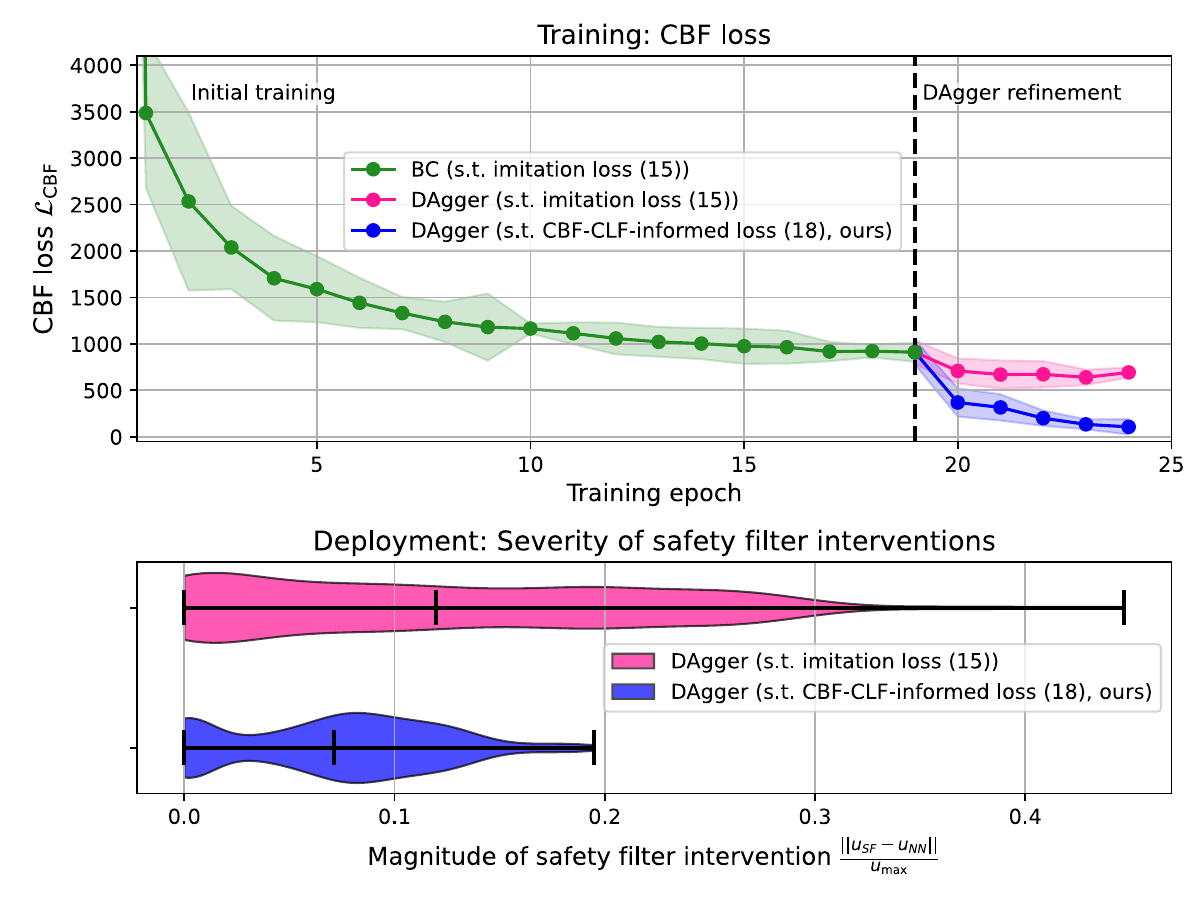
	\caption{Impact of loss function on severity of safety filter interventions during deployment for 50 simulation runs.}
	\label{fig:interventions}
\end{figure}

Figure \ref{fig:interventions} illustrates the influence of the loss function during {DA}\textsc{gger} iterations on the magnitude of safety filter interventions during deployment. On the one hand, solely penalizing the imitation loss \eqref{eq:imit loss} also reduces the CBF loss in both training phases, as the safety constraints are incoporated in the NMPC expert data. On the other hand, refining the initially trained policy by the CBF-CLF-informed loss function \eqref{eq:total loss} clearly minimizes the CBF violations. Crucially, this is reflected in a lower filter activation magnitude, thus proving its benefit within the safety-aware learning approach as our proposed policy internalizes the requirements more closely.  While the conventional {DA}\textsc{gger} subject to imitation loss might achieve a similar CBF loss with increasing number of iterations, modifying the loss function to \eqref{eq:total loss} renders the training significantly more data-efficient.
\begin{table}[t!]
	\begin{center}
		\caption{Comparison of computation time in [ms]. On-device inference for ESP32 is executed without further optimization using \textit{LiteRT}, requiring 246 kB flash memory and 13.6 kB RAM (https://ai.google.dev/edge/litert).}
		\label{tab:runtime}
		\begin{tabular}{|c|c|c|c|c|c|c|}
			\hline
			Policy & (i) & (ii) & (iii) & (iv) & (v, PC) & (v, ESP32)\\
			\hline
			Avg. time & 10.3 & 4.2 & 14.3 & 1.3 & \bfseries 5.8 & \bfseries 28.1\\
			\hline
			Peak time & 39.9 & 16.9 & \bfseries 125.4 & 2.9 & 17.8 & 32.8\\
			\hline
		\end{tabular}
		\vspace{-10pt}
	\end{center}
\end{table}
\vspace{-10pt}

\section{Conclusion} \label{sec: Conclusion}
In this work, we proposed a safety-guaranteed imitation learning framework for spacecraft close proximity control with demonstrated real-time capability on COTS hardware. The presented approach can be tailored to a variety of spacecraft control problems to certify safety of learning-based flight software with onboard autonomy. Future work will focus on extending the concept of operations to investigate safety in subsequent OOS phases such as grasping and detumbling of the target.

\bibliographystyle{IEEEtran}
\bibliography{refs}

\end{document}